\documentclass{article}
\usepackage{arxiv}
\usepackage{cite}
\usepackage{booktabs}
\usepackage[numbers,sort&compress]{natbib}
\usepackage{amsmath,amssymb,amsfonts}
\usepackage{algorithm}
\usepackage{algorithmic}
\usepackage{graphicx}
\usepackage{subfigure}
\usepackage{textcomp}
\usepackage{mathrsfs}
\usepackage{bm}
\usepackage{eucal}
\usepackage{caption}
\usepackage{array}
\usepackage{setspace}
\usepackage{multirow,booktabs,color,soul,threeparttable}
\usepackage{url}

\def\BibTeX{{\rm B\kern-.05em{\sc i\kern-.025em b}\kern-.08em
        T\kern-.1667em\lower.7ex\hbox{E}\kern-.125emX}}

\newcommand{\tabincell}[2]{\begin{tabular}{@{}#1@{}}#2\end{tabular}}
\DeclareMathAlphabet{\mathcal}{OMS}{cmsy}{m}{n}
\DeclareSymbolFont{largesymbols}{OMX}{cmex}{m}{n}

\definecolor{hl}{rgb}{0.75,0.75,0.75}
\sethlcolor{hl}

\begin{document}
\title{Multi-Faceted Representation Learning with Hybrid Architecture for Time Series Classification }

\author{
 Zhenyu Liu \\
  School of Information Management for Law\\
  China University of Political Science and Law\\
  China, Beijing\\
  %% examples of more authors
   \And
 Jian Cheng \\
  China Coal Research Institute CCRI\\
  China, Beijing\\
}

\maketitle

\begin{abstract}
Time series classification problems exist in many fields and have been explored for a couple of decades. However, they still remain challenging, and their solutions need to be further improved for real-world applications in terms of both accuracy and efficiency. In this paper, we propose a hybrid neural architecture, called Self-Attentive Recurrent Convolutional Networks (SARCoN), to learn multi-faceted representations for univariate time series. SARCoN is the synthesis of long short-term memory networks with self-attentive mechanisms and Fully Convolutional Networks, which work in parallel to learn the representations of univariate time series from different perspectives. The component modules of the proposed architecture are trained jointly in an end-to-end manner and they classify the input time series in a cooperative way. Due to its domain-agnostic nature, SARCoN is able to generalize a diversity of domain tasks. Our experimental results show that, compared to the state-of-the-art approaches for time series classification, the proposed architecture can achieve remarkable improvements for a set of univariate time series benchmarks from the UCR repository. Moreover, the self-attention and the global average pooling in the proposed architecture enable visible interpretability by facilitating the identification of the contribution regions of the original time series. An overall analysis confirms that multi-faceted representations of time series aid in capturing deep temporal corrections within complex time series, which is essential for the improvement of time series classification performance. Our work provides a novel angle that deepens the understanding of time series classification, qualifying our proposed model as an ideal choice for real-world applications.

%the temporal dependency of the attribute variables. The visual analyses suggest that the comprehensive

%of various temporal patterns behind the time series data plays an important role for time series classification. Our exploration helps to effectively and efficiently learn the discriminative temporal features of time series data and improve the performance of time series classification-based applications in a variety of fields.
\end{abstract}

\section{Introduction}
\label{sec:intro}

Time series classification (TSC), the task of predicting the predefined class labels for given time series data \cite{chen2013model}, has been explored intensively in the past decades and has been applied in a wide range of fields, such as health care \cite{song2018attend}, fault diagnosis \cite{chen2014cognitive,ChenTRY14}, and financial predictions \cite{liu2017predicting,yang2017granger}. Essentially, any classification problem that takes the temporal ordering into account can be considered a TSC problem. Compared to time-invariant classification problems, TSC problems are characterized by more complex settings, where temporal ordering and correlations, periodicities, and concept drifts interact with each other. Furthermore, time series data are often variable in length and highly dimensional \cite{}. All these issues make TSC problems very challenging for both academic research and real-world applications \cite{chen2015model}.

%In the past decades, many efforts have been conducted to address the TSC challenges \cite{liu2015efficient,jeong2011weighted}.

Generally, conventional machine learning approaches explore TSC problems from two different perspectives: the feature-based approach and the distance-based approach \cite{kate2016using}. In particular, the feature-based approaches, including the Bag-of-Features algorithm (TSBF), Shapelet Transform (ST) \cite{shu2019short}, and Bag-of-SFA-Symbols (BOSS), focus on extracting the most distinguishing features from the original time series data. For instance, TSBF randomly extracts multiple subsequences of different lengths from the original time series and condenses them into patterns to capture the local temporal features \cite{baydogan2013bag}. ST transforms the input time series data into $k$ attributes based on the best $k$ shapelets derived from the original inputs \cite{hills2014classification}. BOSS transforms the input time series data into the corresponding histogram using the symbolic Fourier approximation \cite{schafer2015boss}. In contrast, the distance-based approaches predict the class labels of the test samples based on their distances to the training samples. For example, the Euclidean distance (ED) measures the distance between two time series of equal lengths \cite{lines2015time}. The Dynamic Time Warping (DTW) distance warps time series of different lengths to be the same length and defines an alignment path with the minimum cumulative distance in the distance matrix \cite{keogh2001derivative}. When coupled with the $k$ nearest neighbor ($k$-NN) classifier, the DTW approach provides a strong baseline.

%Feature-based approaches adopt various means to extract a set of features that can reflect temporal patterns hidden in time series .

%, the distance measurement between the informed data are more effective to distinguish the different classes.
% are more distinguishing over the extracted feature variables,

%intend to gather together and different classes disperse spatially discriminately across the space. In other words, the class of an unlabeled data instance can be estimated by the distance measured with the data instance of the known classes in a feature space. Therefore, how to effectively design/extract most distinguishing representations of time series data and how to accurately measure the distance between instance data are two main aspects of traditional TSC approaches.

%Thereby, a distance-based classifier can distinguish two time series based on their histograms. Although experimental studies have indicated that feature-based methods present advantages in classification accuracy, they suffer from heavy-weight feature extraction and feature engineering.

Notwithstanding differences in perspectives, these approaches share certain characteristics in common. Essentially, the original temporal signals are transformed into a new feature space where different classes can be distinguished from each other more effectively. Furthermore, the finding that combining a number of $k$-NN classifiers with various distance measures can outperform any individual $k$-NN classifier \cite{lines2015time} has motivated further work on this topic. Numerous studies have shown that combining different classifiers over a variety of feature spaces, rather than different classifiers over the same feature space, can significantly improve the performance of TSC. This knowledge led to a variety of ensemble approaches for TSC, such as the Elastic Ensemble (EE), Shapelet Ensemble (SE), Collective Of Transformation-based Ensembles (COTE), and its augmented hierarchical voting version (HIVE-COTE). The EE approach integrates $1$-NN classifiers based on 11 elastic distance measures with a voting scheme \cite{lines2015time}, and SE adopts a heterogeneous ensemble strategy with transformed shapelets to produce the classifiers \cite{bagnall2015time}. In addition, COTE and HIVE-COTE, which combine 35 different classifiers for different time series representations, have shown significant improvement in classification accuracy when evaluated with the datasets from the UCR repository. They are now considered to be the state-of-the-art for solving TSC problems \cite{bagnall2015time, fawaz2019deep}.

The success of the ensemble approaches is based on the holistic representations of the time series data. Hence, the performance of ensemble models can be improved by continuously increasing their sizes. However, as the size of an ensemble model increases, its computational complexity increases as well. As a result, when attempting to solve ``big data’’ problems, the massive ensemble models are infeasible to train and run \cite{bagnall2017great}. For instance, in COTE, one of the 35 classifiers is ST, the time complexity of which is $O(n^2 \cdot l^4)$, where $n$ is the number of time series data and $l$ is the length of the time series. Additionally, the fact that the ensemble of a large number of classifiers is constructed over different feature spaces makes the models uninterpretable for the domain experts who design the features. Moreover, as the manually constructed features of conventional TSC models are highly subject to human domain knowledge, it is usually difficult for the models based on such features to be transferred to new domains. These limitations of conventional approaches hinder further progress for TSC. Therefore, a new approach that is domain agnostic and can learn holistic representations from time series without laborious manual effort is required.

%However, makes it a hugely large ensemble model . besides the heavy hand feature engineering, ensemble approaches suffer from the problem of huge computation, and they are considered to be limited if not impractical  in the real-time setting for real-world applications. Again, the ensemble-based approaches also suffer from unbearable data preprocessing and the heavy hand feature engineering.

%However, the drawback of this paradigm is that a single classifier is lack of effective adaptivity across domains with different temporal structures (i.e. when faced with new scenarios, the classifier has to be retrained on the new datasets), thus its performance dropping significantly in complex situations.

Deep neural networks (DNNs), which are used for learning hierarchical representations from data, provide such a promising approach. In the past decade, DNNs have been widely applied and have achieved impressive success in a variety of pattern recognition tasks  \cite{lecun2015deep}. There are many different architectures of DNNs, such as Convolutional Neural Networks (ConvNets) and Recurrent Neural Networks (RNNs), each of which features a distinctive representation learning capacity. Among the different DNN architectures, RNNs, especially the long short-term memory (LSTM) ones \cite{hochreiter1997long}, were purposely designed to address sequential prediction problems, such as natural language processing (NLP) and speech recognition \cite{lecun2015deep}. However, as RNNs suffer from overfitting problems and require large amounts of computer memory and time for training, they are rarely applied to solve TSC problems. In contrast, 1D filters of ConvNets, while originally designed for modeling spatially invariant features with 2D filters, have been applied to capture the information changes in the temporal dimension of time series. Such temporal ConvNets exhibit superior robustness in learning time-invariant features in raw input time series and require much less time and memory to train. Therefore, various kinds of temporal ConvNets have been proposed for TSC, such as multi-scale convolutional neural networks (MCNNs) \cite{cui2016multi}, fully convolutional networks (FCNs) \cite{wang2017time}, and residual networks (ResNets) \cite{wang2017time}.

In particular, FCNs have demonstrated a superior performance with relatively few parameters, which is important for TSC because of the scarcity of carefully labeled time series data. Thus, FCNs are currently considered a very strong DNN baseline for solving TSC problems \cite{wang2017time}.
However, it is difficult for any existing neural architecture to completely capture the temporal correlations in time series due to the complexities involved. For instance, FCNs have the capability of extracting locally temporal features at different levels, but they cannot capture the temporal dependencies over long durations due to the limited size of the temporal convolution filters. In contrast, RNNs can learn long-term dependencies in sequential data, but they offer more support for the analysis of recent information than they do for older information, which inevitably leads to the distortion of temporal patterns.

%some intrinsic limitations to accomplish high performance in robustness and generalization.

%difficulties of high dimensional, auto-correlated and diversity in distribution of different domains, there is still a much-needed demand to improve the generalization capability of DNN based approaches \cite{serra2018towards,fawaz2019deep}.

%some well-known deep learning-based approaches have achieved competitive performance for TSC.
%The prominent characteristics
%to take the auto-correlated structure information within time series into consideration properly because of .

In this study, we propose a hybrid neural architecture, called Self-Attentive Recurrent Convolutional Networks (SARCoN), to model the temporal correlation in time series by integrally learning representations of different aspects of time series using different architectures. Our work is based on the rationale of the current state-of-the-art ensemble approaches, which learn holistic representations of times series by synthesizing different individual classifiers. In SARCoN, FCNs are used as the submodule to capture the local temporally invariant features at multiple levels, and LSTM with a self-attention mechanism is applied as the submodule to learn long-distance dependencies in the input time series. Moreover, rather than simply concatenating the pretrained neural networks in the conventional ensemble manner, all the submodules of SARCoN are jointly trained in a pure end-to-end manner. When predicting the unseen time series, the different submodules of SARCoN work in parallel to extract constituent representations, which are then combined into the multi-faceted representations of the input time series. The resulting multi-faceted representations can then be fed into a softmax layer for classification.

%validate our posture that ...

%avoiding increase the depth or the width of the network,

%Inspired by a deep and wide model in \cite{cheng2016wide},

The contributions of our work can be summarized as follows:
\begin{itemize}
\item We propose a hybrid end-to-end architecture for TSC, called SARCoN, that can learn multi-faceted representations from time series. In the proposed approach, the ensemble of the full angle representations, including both the locally temporal features and the long-distance temporal dependencies, are validated in that they exhibit a superior performance over the current state-of-the-art approaches.

\item We conduct a series of well-designed experiments on the UCR benchmark datasets in an attempt to interpret the mechanism of classification for TSC. With the aid of the self-attention and global average pooling layers, we present an overall analysis, both theoretically and visually, of the determining factors for the classification decision. This contributes to a deeper understanding of temporal correlations in time series.

\end{itemize}

The contents of this paper are organized as follows. Section \ref{sec:rw} reviews related work on DNN approaches for TSC. Section \ref{sec:SARCoNs} details the overall architecture of our proposed approach. Our experimental results and related analyses are presented in Section \ref{sec:expers}. Finally, Section \ref{sec:conclu} gives the summary of this paper and the discussion of future work.

\section{Related Work}
\label{sec:rw}

In this paper, we focus on exploring the synthesis of different types of neural structures, namely ConvNets, RNNs, and attention mechanisms, to learn holistic representations of time series. Here, we give a brief introduction to the background knowledge on ConvNets, RNNs, attention mechanisms, ensemble learning and sequential representation learning.

\subsection{Convolutional Networks}

ConvNets have achieved significant success in a wide range of domains, including computer vision, natural language processing, and speech recognition \cite{lecun2015deep,ioffe2015batch}. In ConvNets, convolutional filters are designed to detect local conjunction patterns, and the pooling operations are used to merge semantically similar features \cite{lecun2015deep}. The hierarchy of the convolutional and pooling layers are adept at learning compositional semantic features, in which the higher level features are composed of the lower level ones.

For TSC, ConvNets usually employ 1D convolutional filters to learn invariant features along the time dimension, a process called temporal convolution \cite{lea2016temporal}. There are several temporal ConvNets-based models that have exhibited a performance that is competitive with the conventional state-of-the-art by using different methods. For instance, MCNN extracts ad hoc subsequences from the original time series using downsampling, skip sampling, and sliding windows to augment the training dataset for multiple scale settings \cite{cui2016multi,gong2016model}. However, as MCNN relies heavily on the domain knowledge for data preprocessing and feature selection, it performs generalization poorly and is incompatible with the end-to-end architecture \cite{fawaz2019deep}.

The exploration of large (i.e., deep and/or wide) structures is informative for DNN structuring. To the best of our knowledge, ResNets are currently the deepest models for TSC, with an architecture \cite{fawaz2019deep} featuring 11 layers. Such a deep structure enables ResNets to gain the powerful ability to learn complex patterns from time series. However, it also leads to the susceptibility to overfit because of the sparsity of the time series training data. As MCNN and ResNets are not structured in accordance with our objective of developing a model that is domain agnostic and robust for sparse training data, these two structures are not considered in this study.

FCNs have a structure similar to ResNets, but are more lightweight. As ResNets, FCNs are end-to-end models for TSC. The number of parameters of each layer in FCNs is kept constant throughout the convolutions, which means that FCNs are transferable to different datasets. In FCNs, temporal convolutions are used to learn temporally invariant features, and global average pooling is adopted to capture the global compositional semantic features. As previous studies demonstrate that FCNs achieve a similar or better performance than conventional state-of-the-art COTE and the more complicated DNN rival ResNets \cite{wang2017time}, we adopt FCNs as the DNN benchmark in our experiments.

In the process of combining different types of DNN structures, which is designed to enhance the overall performance, ConvNets is the main component. It is usually combined with other types of structures, such as a gated recurrent unit (GRU) \cite{lin2017gcrnn} and an attention mechanism \cite{serra2018towards}. In these hybrid models, the sequential time steps of the time series are viewed as multidimensional data with one time step to reduce the computation overheads. However, such processing simplifies the complex temporal correlations in time series, thus still leading to the loss of long-distance temporal dependencies.

\subsection{Recurrent Networks}

RNNs are another family of neural structures, and they also represent the powerful dynamic systems modeling of long-term dependencies in sequential data \cite{lecun2015deep}. When processing sequential data, an RNN takes one element in the sequence at each time step and merges the current input with the history information of the past time steps to learn the inherent patterns. However, due to these recurrent operations, RNNs suffer from the gradient escalating or vanishing during training. Due to advances in RNN structure, the difficulties in training RNNs can be addressed by many recently developed types of RNN variants, such as LSTM, GRUs, and echo state networks (ESNs) \cite{liu2017predicting,tang2014learning}. Although RNN-based models are widely used for sequential predictions, they are rarely used for TSC, mainly because of the relatively high computation complexity.

Given the limitations of RNNs for solving TSC problems, an ESN-based approach, named time warping invariant echo state network (TWIESN), was proposed as the first pure RNN for solving TSC problems \cite{tanisaro2016time}. However, the experimental studies conducted by \cite{fawaz2019deep} showed that the RNN-based approaches are still in their initial stages of development, failing to provide performances comparable to the state-of-the-art.

In this study, we enhance FCNs by leveraging the LSTM structure to learn long-distance temporal dependencies in time series. Currently, most applications of LSTM for sequential predictions adopt the two-way processing approach, in which the input sequences are processed from the front end and back end simultaneously. However, in this work, we use the one-way LSTM for processing the input time series to reconcile the one-way temporal ordering of most natural time series data.

\subsection{Attention Mechanisms}
In recent years, the different kinds of attention mechanisms have become the integral ingredients for DNNs due to their flexible power in modeling dependencies and parallelized computation \cite{vaswani2017attention}. In most cases, attention mechanisms are used in conjunction with RNNs, as they allow the modeling of dependencies regardless of the distances in the input sequences, which mitigates the difficulties of using RNNs for sequential computation subject to memory constraints. With their significant advantage in modeling dependencies, attention mechanisms have also been employed for TSC. For instance, the sequence-to-sequence model with attention (S2SwA) \cite{tang2016sequence} demonstrates the effectiveness of the sequence-to-sequence framework, which incorporates the LSTM encoder and decoder to solve TSC problems.

Self-attention, also called intra-attention, is a special kind of attention mechanism that relates different positions of a single sequence to compute a new representation of the sequence. Self-attention has been widely used in a variety of tasks, including image generation \cite{zhang2018self} and text classification \cite{liu2020hierarchical}. Based on the work in \cite{lin2017structured}, which utilized self-attention to learn multi-faceted temporal patterns, in this study we develop a self-attentive LSTM to augment representation learning in FCNs.

\begin{figure*}[t]
    \centering
    \includegraphics[width=16cm]{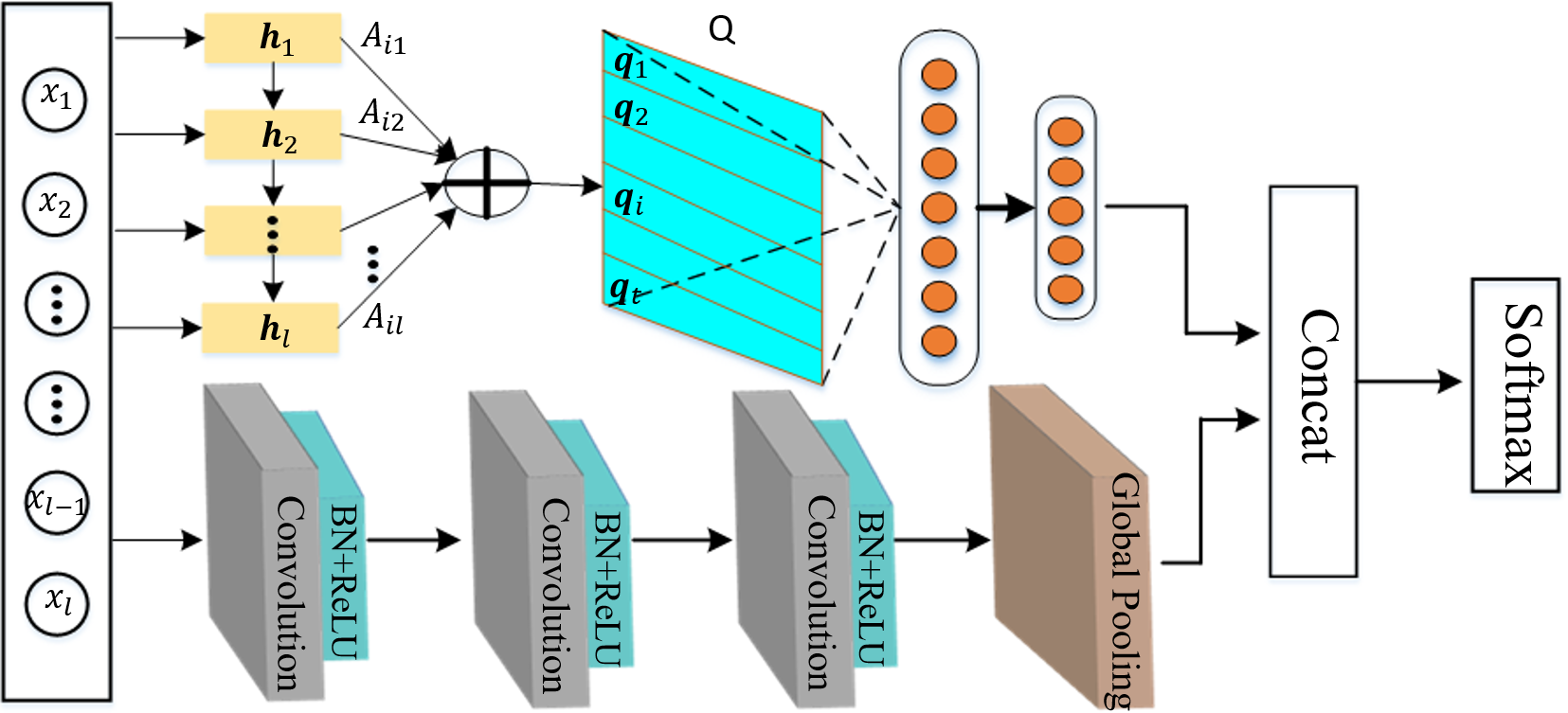}
    \caption{The overall structure of our proposed method. The upper part is the Structured Self-Attentive (SSA) network block, and the lower part is the Fully Convolutional Network (FCN) block. The sequence embedding Q is computed as multiple weighted sums of hidden states from an LSTM $\left(\bm{h}_1,\bm{h}_2,...,\bm{h}_l\right)$. The variable $A$ represents the summation weights matrix. Features extracted from the two blocks are passed into a softmax layer for predicting the probability distribution across category values.}
    \label{fig3:SSA-FCN}
\end{figure*}

\subsection{Ensemble Learning}

Ensembles learning is a widely used and effective technique in the machine learning community, which combines a set of accurate yet diverse classifiers for better generalization. Ensemble learning is successful due to the concept of ``diversity'' \cite{chen2008diversity,bian2019does,chen2007evolutionary}. There are several widely-used ensemble algorithms, such as Bagging, Boosting, random forests, negative correlation learning (NCL)\ cite{chen2007evolutionary,chen2009regularized}, etc. NCL tries to generate a compact ensemble that generates a negative correlated ensembles \cite{chen2010multiobjective,chen2007evolutionary,chen09}. Ensemble learning usually consumes a lot of resources for better generalization. To reduce the computational complexity, the ensemble pruning algorithms are often employed \cite{bian2019ensemble,chen2009predictive,chen06probabilistic}.

\subsection{Sequential Representation learning}
In the sequential processing model, there are a number of representation learning methods, such as fisher kernel and model in the learning space \cite{li2016sequential}. In the learning in the model space, it transforms the original temporal series to an echo state network (ESN), and calculates the `distance' between ESNs  \cite{ChenTRY14,chen2013model}. Therefore, the distance based learning algorithms could be employed in the ESN space \cite{gong2016model}. Chen et al. \cite{chen2015model} investigated the trade-off between the representation and discrimination abilities. Gong et al. proposed the multi-objective version for learning in the model space \cite{gong2018multiobjective}.

\section{Self-Attentive Recurrent Convolutional Networks}
\label{sec:SARCoNs}

In this section, we first clarify our notation and give a definitive description of the TSC problems. Then, we give the details of the architecture of SARCoN, in which self-attentive LSTM and FCNs are introduced in details as two feature extraction \cite{wu2019accurate,jiang2019probabilistic} submodules for the univariate time series. At the end of this section, we introduce the loss function, optimizer, and evaluation criteria we used to train SARCoN.

\subsection{Problem Formalization and Notations}

We denote vectors with bold lowercase letters and matrices with uppercase letters. The TSC problems in this paper are formulated as follows.

\begin{itemize}
    \item A univariate time series is a sequence of real-value data points listed in time order and separated by uniform time intervals.

    \item A labeled time series dataset with $N$ instances is denoted as $X{\rm{ = }}\left\{ {\left( {{\bm{x}^i},{y_i}} \right)} \right\}_{i = 1}^N$, where $(\bm{x}^i,y_i)$ represents an instance pair with $l$ observations $(x_1, x_2, ..., x_l)$ and class label $y_i$ with $C$ possible values.

    \item The objective of the TSC problem is to learn a classifying function that predicts the probability distribution over the labels for the given inputs.

\end{itemize}

\subsection{Overall Architecture}

The overall network architecture of the proposed SARCoN is illustrated in Figure \ref{fig3:SSA-FCN}. In this architecture, the self-attentive LSTMs and FCNs are connected in parallel to learn representations from various perspectives.

The top half of Figure \ref{fig3:SSA-FCN} illustrates the structure of the unidirectional self-attentive LSTMs, which learn the representations of the long-distance semantic information in the input series. The bottom half of Figure \ref{fig3:SSA-FCN} illustrates the structure of the FCNs, which learn the time-invariant temporal features in the input series. These two submodules augment each other and provide a complementary multi-faceted representation of the input series. After obtaining the embedded representations from these two modules, we concatenated them in the last layer of the self-attentive LSTM and FCN modules. Subsequently, the merged features are passed into a softmax layer for the final time series classification. The overall networks are then jointly trained in an end-to-end manner.

\subsection{Self-Attentive LSTM}

In the self-attentive LSTM submodule, an LSTM cell maintains a hidden state vector, $\bm{h}_t$, and a memory cell, $\bm{c}_t$. The latter is constantly updated at each time step. The success of LSTM lies in the design of the three gates, which are the input gate ($\bm{i}_t$), the output gate ($\bm{o}_t$), and the forget gate ($\bm{f}_t$). They are able to determine what information should be preserved or removed in the internal state. The information flow transferred inside the LSTM at each time step is

\begin{equation}
\begin{array}{l}
    \displaystyle {\bm{i}_t} = \sigma ({W^i}{\bm{h}_{t - 1}} + {V^i}{\bm{x}_t} + {\bm{b}_i}), \\
    \displaystyle {\bm{f}_t} = \sigma ({W^f}{\bm{h}_{t - 1}} + {V^f}{\bm{x}_t} + {\bm{b}_f}), \\
    \displaystyle {\bm{o}_t} = \sigma ({W^o}{\bm{h}_{t - 1}} + {V^o}{\bm{x}_t} + {\bm{b}_o}), \\
    \displaystyle {\bm{g}_t} = \phi ({W^g}{\bm{h}_{t - 1}} + {V^g}{\bm{x}_t} + {\bm{b}_g}), \\
    \displaystyle {\bm{c}_t} = {\bm{f}_t} \odot {\bm{h}_{t - 1}} + {\bm{i}_t} \odot {\bm{g}_t}, \\
    \displaystyle {\bm{h}_t} = {\bm{o}_t} \odot \phi ({\bm{c}_t}),
\end{array}
\end{equation}
where $\odot$ denotes elementwise multiplication; $\sigma \left(  \cdot  \right)$ represents the logistic sigmoid function; $\phi \left(  \cdot  \right)$ represents the tanh function; $W^g, W^i, W^o,$ and $W^f$ are recurrent weight matrices; and $V^g, V^i, V^o,$ and $V^f$ are projection matrices.

When a time series $\bm{x} = \left(x_1,x_2,...,x_l\right)$ is fed into a unidirectional LSTM, a hidden state sequence $H = \left(\bm{h}_1,\bm{h}_2,...,\bm{h}_l\right)$ (the size of which is $l$-by-$n$) is obtained, where $\bm{h}_i \in R^n$ is the state vector at the $i$-th time step. The self-attention mechanism is applied to compute a linear combination of the $l$ LSTM hidden states in $H$. An attention vector $\bm{a}$ is calculated by

\begin{equation}
\bm{a} = softmax ({\bm{w}^2}\tanh ({W^1}{H^T})),
\label{eq:weight vector}
\end{equation}
where $W^1 \in R^{s \times n}$ indicates weight parameters and $\bm{w}^2 \in R^{s}$ is a vector of parameters. Here, $s$ is an assignable hyperparameter. A vector representation $\bm{q}$ of the input time series can be obtained by a weighted sum over the LSTM hidden states $H$. In order to better capture the rich information in the sequence, multiple $q$'s that focus on various patterns of the sequence are needed. Accordingly, $\bm{w}^2$ is extended into a matrix $W^2$ with a $t$-by-$s$ shape. The resulting attention matrix $A$ is

\begin{equation}
A = softmax ({W^2}\tanh ({W^1}{H^T})).
\end{equation}
Finally, $t$ represented vectors are computed to obtain the sequence embedding $Q$ by
\begin{equation}
Q=AH.
\label{eq:se}
\end{equation}

This self-attentive LSTM acts as one of the feature extractors in SARCoN to learn multi-faceted representations from the input sequences with variable lengths.

\subsection{Fully Convolutional Networks}
Similar to the self-attentive LSTM submodule, the FCN submodule functions as another feature extractor. The inputs to the FCN are time series signals. Let ${S_t} \in {R^{{V_0}}}$ indicate the input vector with dimension $V_0$ at time step $t$, where $0 < t \le T$ and $T$ is the length of each sequence. The true label for each frame is given by $y_t \in \left\{{1, 2,..., C}\right\}$, where $C$ represents a finite number of categories. For each of the $L$ convolutional layers, a set of 1D filters are employed to capture how the input signals evolve over the course of an action. Let tensor ${W^\ell} \in {R^{{V_\ell} \times d \times {V_{\ell - 1}}}}$ and biases $ {b_\ell} \in {R^{{V_\ell}}}$ be filter parameters for each layer, where $\ell \in \left\{{1, 2,..., L}\right\} $ represents the layer index and the filter duration is denoted by $d$. For the $\ell$-th layer, the $i$-th component of the unnormalized activation $\hat{E}_t^{\ell} \in R^{V_\ell}$ is generated using the (normalized) activation matrix ${E}^{\ell-1} \in {R^{V_{\ell-1} \times T_{\ell-1}}}$ from the previous layer at each time $t$ via

\begin{equation}
\hat{E}_{i,t}^{(\ell)} = f\left(\sum\limits_{t^\prime = 1}^{d} \langle W_{i,t^\prime,.}^{(\ell)}, E_{.,t+d-t^\prime}^{(\ell - 1)} \rangle + b_i^{(\ell)} \right),
\end{equation}
where $f(\cdot)$ is a Rectified Linear Unit (ReLU).

\begin{figure}[t]
    \centering
    \includegraphics[width=8.5cm]{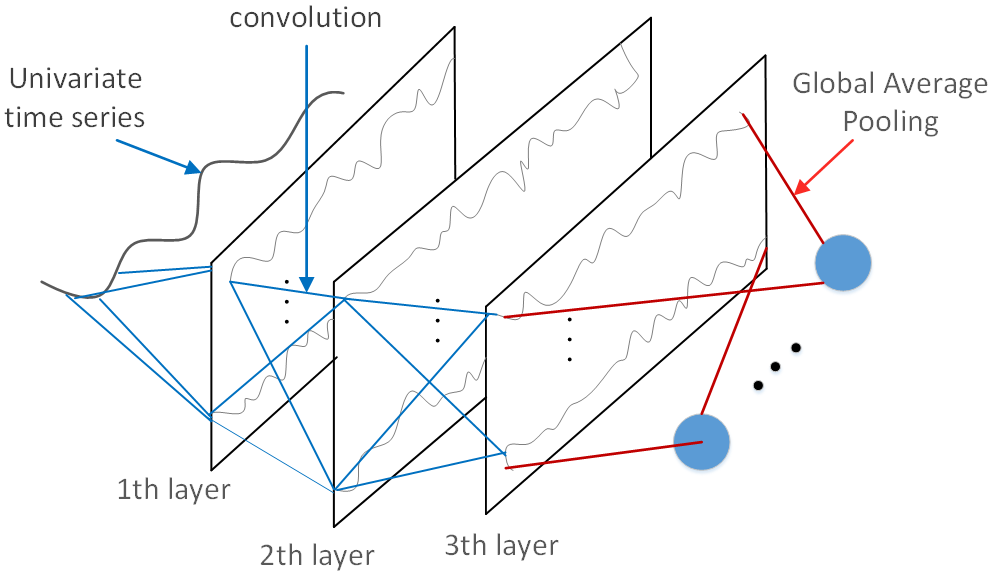}
    \caption{The architecture of a FCN submodule adopted in the proposed model. The input to the FCN block is a univariate time series of variable length, and three basic blocks are employed to extract time invariant features. The features extracted from the third layer are fed into a global average pooling layer.}
    \label{fig2:FCN}
\end{figure}

The basic block of the FCN submodule consists of the temporal convolutional network (TCN) followed by a batch normalization layer \cite{ioffe2015batch} and a ReLU activation layer. In this architecture, a TCN is employed as a convolutional layer. Batch normalization can accelerate the training process and enhance the generalization. The formula is

\begin{equation}
\begin{array}{l}
\displaystyle  C = W \otimes \bm{x} + \bm{c},\\
\displaystyle  B = BatchNormalization(C),\\
\displaystyle  H = ReLU(B),
\end{array}
\end{equation}
where $\otimes$ indicates the convolution operator. The FCN stacks three such basic blocks, and a global average pooling layer averages each representation from the third layer, which significantly reduces the number of weights. Figure \ref{fig2:FCN} shows a brief illustration of the FCN submodule.

\subsection{Loss Function and Optimization}
In this paper, we solve both the binary and multi-class time series tasks. Accordingly, the cross entropy loss function is defined as

\begin{equation}
loss =  - \sum\limits_{n = 1}^N {\sum\limits_{c = 1}^C {y_{nc}^t\log \left( {y_{nc}^p} \right)} },
\label{eq:cross entropy}
\end{equation}
where $y_{nc}^t$ denotes the true probability that the $n$-th sample belongs to class $c$ and $y_{nc}^p$ is the corresponding predicted probability. Our proposed method is trained with the Adam optimizer \cite{kingma2014adam} (which has proven to be a very effective optimizer) using an initial learning rate of $1e^{-3}$. If the validation score stops improving after every 100 epochs, the learning rate is reduced by a factor of $1 / {\sqrt[3]{2}}$ until the final learning rate falls below $1e^{-4}$.

\section{Experiments and Analyses}
\label{sec:expers}

In this section, we first give a description of the experimental settings and the datasets employed. Then we evaluate the performance of the proposed algorithm using 24 time series benchmarks and compare its accuracy to other state-of-the-art algorithms. The visualization of contributing regions produced by the two submodules aids in the understanding of the role of each submodule. At the end of this section, we visualize the features from the stitching layer to demonstrate our model’s capacity for discriminative feature extraction \cite{jiang2019joint,he2015robust}.

\subsection{Time Series Benchmark}

As univariate time series provide a reasonably simple starting point to explore temporal signals, we conducted experiments on 24 time series benchmarks from the UCR repository \cite{chen2015ucr} to demonstrate the effectiveness of our proposed approach. The datasets with different lengths were generated from various fields. The minimum and maximum number of classes were 2 and 60, respectively. The sizes of the training sets varied from 55 to 3636, and most of the sequence lengths were longer than 500. All default datasets from the UCR repository were preprocessed into training and testing sets without other additional preprocessing because the data have almost zero mean and unit variance. For more information about the datasets, refer to Table \ref{tab:datasets}.

\begin{table}
    \centering
    %\caption{Detailed information on the 24 time series benchmarks.}
    \caption{The detailed information of 24 time series benchmarks}
    \label{tab:datasets}
    \begin{tabular}{lcccc}
        \toprule
        Datasets& Train& Test& Length& Classes\\
        \midrule
        Car& 60& 60& 577& 4\\
        Computers& 250& 250& 720& 2\\
        FordA& 1320& 3601& 500& 2\\
        FordB& 3636& 810&  500& 2\\
        HandOutlines& 1000& 370& 2709& 2\\
        Haptics& 155& 308& 1092& 5\\
        Herring& 64& 64& 512& 2\\
        InlineSkate& 100& 550& 1882& 7\\
        LargeKitApp& 375& 375& 720& 3\\
        Lighting2& 60& 61& 637& 2\\
        Mallat& 2345& 55& 1024& 8\\
        NonECG1& 1800& 1965& 750& 42\\
        NonECG2& 1800& 1965& 750& 42\\
        OliveOil& 30& 30& 570& 4\\
        Phoneme& 1896& 214& 1024& 39\\
        RefDevices& 375& 375& 720& 2\\
        ScreenType& 375& 375& 720& 2\\
        ShapesAll& 600& 600& 512& 60\\
        SmallKitApp& 375& 375& 720& 3\\
        StarLCurves& 8236& 1000& 1024& 3\\
        Strawberry& 613& 370& 235& 2\\
        uWavGestAll& 3582& 896& 945& 8\\
        Worms& 181& 77& 900& 5\\
        WormsTwoClass& 181& 77& 900& 2\\
        \bottomrule
    \end{tabular}
\end{table}

\subsection{Comparative Baselines}
The comparative baselines in our experiments covered almost all the widely adopted conventional approaches, including ED \cite{gong2018sequential}, $\textup{DTW}_F$ \cite{kate2016using}, TSBF \cite{baydogan2013bag}, ST \cite{hills2014classification}, and BOSS \cite{schafer2015boss}. In particular,
our findings were empirically compared to conventional state-of-the-art approaches, such as EE \cite{lines2015time} and COTE \cite{bagnall2015time}. Additionally, the performance comparison was conducted between our proposed SARCoN and the current DNN golden baseline FCN \cite{wang2017time}. The experimental settings of these algorithms can be found in \cite{wang2017time}.

\subsection{Evaluation Metrics}
For time series classification, accuracy is a common evaluation indicator. In addition, we adopted several other evaluation criteria, such as the number of wins, rank-based statistics, and Mean Per Class Error (MPCE) \cite{wang2017time}.

The number of wins is the number of datasets on which the model performs best. The rank-based statistics include the Arithmetic Rank (AR) and the Geometric Rank (GR). The AR and GR are the arithmetic mean and the geometric mean, respectively, of the ranks of the datasets. To consider the factor of the number of classes, the MPCE was employed as a robust metric to evaluate the performance of the compared models across multiple datasets. The MPCE is the simple arithmetic mean of the Per Class Error (PCE), and it is defined as
\begin{equation}
PC{E_m} = \frac{{Erro{r_m}}}{{{C_m}}},
\end{equation}
\begin{equation}
    MPC{E_i} = \frac{1}{M}\sum {PC{E_m}},
\end{equation}
where $m$ is the $m$-th dataset out of a total of $M$ datasets, $i$ denotes the $i$-th compared model, $Error_m$ is the error rate of the $i$-th model for the $m$-th dataset, $C_m$ is the number of unique categories for the $m$-th dataset, and $MPCE_i$ is the expected error of the PCE across all the datasets.

\subsection{Experimental Setup}
In the experiments, the FCN block and the Structured Self-Attentive (SSA) block remained constant. The specific parameters of the three basic blocks in the FCN were as follows: the first convolutional layer contained 128 filters of length 8, the second convolutional layer was composed of 256 filters of length 5, and the third convolutional layer was composed of 128 filters of length 3. The stride of the convolution kernel on the three convolutional layers was 1, and the number of LSTM cells in the SSA block was 128. The hyperparameter $s$ was set to 350, and 30 weighted sums ($t$) were computed by multiplying the attention matrix and the sequence matrix together. The number of neurons in the first and second fully connected layers were 2000 and 128, respectively. A high dropout rate of 0.8 was used between the two fully connected layers to reduce overfitting. The number of epochs was usually set to 1500, but sometimes it was increased because the algorithm required more time to converge for certain datasets. The batch size was optimally selected from $\{{128, 64, 32}\}$, and a class weighting scheme \cite{king2001logistic} was adopted for the class imbalance that appeared in specific datasets.

\begin{table*}[!tbp]
    \centering
    %\caption{Comparison between classification accuracy of our proposed model and other algorithms for 24 time series benchmarks. The dataset with the highest accuracy for each model is marked with a shadow.}
    \caption{Classification accuracy of our proposed model and other comparison algorithms on 24 time series benchmarks. The highest accuracy of each dataset for listed models has been marked with shadow}
    \label{tab:result}%添加标题 设置标签
    \begin{tabular}{lccccccccc}
        \toprule
        Datasets& ED& $\textup{DTW}_F$& TSBF& ST& BOSS& EE& COTE& FCN& SARCoN\\
        \midrule
        %Adiac& 0.611& 0.605& 0.727& 0.768& 0.749& 0.665& 0.81& 0.857& \hl{0.872}\\
        %ArrowHead& 0.8& 0.776& 0.801& 0.851& 0.875& 0.86& 0.877& 0.88& \hl{0.897}\\
        %Beef& 0.667& 0.546& 0.554& 0.736& 0.615& 0.532& \hl{0.867}& 0.75& 0.833\\
        %BeetleFly& 0.75& 0.853& 0.799& 0.875& 0.949& 0.823& 0.921& \hl{0.95}& \hl{0.95}\\
        %BirdChicken& 0.55& 0.865& 0.902& 0.927& 0.984& 0.848& 0.941& \hl{1}& \hl{1}\\
        Car& 0.733& 0.851& 0.795& 0.902& 0.855& 0.799& 0.899& 0.917& \hl{0.95}\\
        %CBF& 0.852&    0.979& 0.977& 0.986& 0.998& 0.993& 0.998& \hl{1}& 0.997\\
        %ChlorineConcentration& 0.65& 0.658&    0.683& 0.682& 0.66& 0.659& 0.736& 0.843& \hl{0.998}\\
        %CinCECGtorso& 0.897& 0.714& 0.716& 0.918& 0.9& 0.946& \hl{0.983}& 0.813& 0.726\\
        %Coffee& \hl{1}& 0.973& 0.982& 0.995& 0.989& 0.989& \hl{1}& \hl{1}& \hl{1}\\
        Computers& 0.576& 0.659& 0.765& 0.785& 0.802& 0.732& 0.77& 0.848& \hl{0.868}\\
        %CircketX& 0.577& 0.769& 0.731& 0.777& 0.764& 0.801& 0.814& 0.815& \hl{0.818}\\
        %CircketY& 0.567& 0.756& 0.728& 0.762& 0.749& 0.794& \hl{0.815}& 0.792& 0.805\\
        %CircketZ& 0.587& 0.785& 0.738& 0.798& 0.776& 0.804& 0.827& 0.813& \hl{0.844}\\
        %DiatomSizeReduction& 0.935& 0.942& 0.89& 0.911& 0.939& 0.946& 0.925& 0.93& \hl{0.951}\\
        %DistalPhalanxOAG& 0.626& 0.796& 0.816& 0.829& 0.815& 0.768& 0.805& 0.835& \hl{0.86}\\
        %DistalPhalanxOC&  0.717& 0.76& 0.812& 0.819& 0.814& 0.768& 0.821& 0.812& \hl{0.83}\\
        %DistalPhalanxTW& 0.633& 0.658& 0.69& 0.69& 0.673& 0.654& 0.693& 0.79& \hl{0.818}\\
        %Earthquakes& 0.713& 0.747& 0.747& 0.737& 0.746& 0.735& 0.747& 0.801& \hl{0.826}\\
        %ECG200& 0.88& 0.819& 0.847&    0.84& 0.89& 0.881& 0.873& 0.9& \hl{0.92}\\
        %ECG5000& 0.925& 0.94& 0.938& 0.943& 0.94& 0.939& 0.946& 0.941& \hl{0.947}\\
        %ECGFiveDays& 0.798& 0.907& 0.849& 0.955& 0.983& 0.847& 0.986& 0.985& \hl{0.999}\\
        %ElectricDevices& 0.552& 0.874& 0.808& \hl{0.895}& 0.8& 0.831& 0.883& 0.723& 0.786\\
        %FaceAll& 0.715& 0.963& 0.942& 0.968& 0.974& 0.976& \hl{0.99}& 0.929& 0.944\\
        %FaceFour& 0.785& 0.909& 0.862& 0.794& \hl{0.996}& 0.879& 0.85& 0.932& 0.932\\
        %FacesUCR& 0.769& 0.889& 0.849& 0.909& \hl{0.951}& 0.948& 0.967& 0.948& 0.939\\
        %FiftyWords& 0.632& 0.748& 0.744& 0.713& 0.702& \hl{0.821}& 0.801& 0.679& 0.791\\
        %Fish& 0.783& 0.931& 0.913& 0.947& 0.969& 0.913& 0.962& 0.971& \hl{0.977}\\
        FordA& 0.665& 0.884& 0.831& \hl{0.965}& 0.92& 0.751& 0.955& 0.906& 0.922\\
        FordB& 0.616& 0.884& 0.751& 0.915& 0.911& 0.757& \hl{0.929}& 0.883& 0.92\\
        %Gun Point& 0.913& 0.964& 0.965& 0.999& 0.994& 0.974& 0.992& \hl{1}& \hl{1}\\
        %Ham&  0.6& 0.795& 0.711& 0.808& \hl{0.836}& 0.763& 0.805& 0.762& 0.791\\
        HandOutlines& 0.863& 0.915& 0.879& \hl{0.924}& 0.903& 0.88& 0.894& 0.776& 0.864\\
        Haptics& 0.371& 0.464& 0.463& 0.512& 0.459& 0.451& 0.517& \hl{0.551}& 0.526\\
        Herring& 0.516& 0.609& 0.59& 0.653& 0.605& 0.566& 0.632& 0.703& \hl{0.734}\\
        InlineSkate& 0.343& 0.382& 0.377& 0.393& \hl{0.503}& 0.476& 0.526& 0.411& 0.475\\
        %InsectWingbeatSound& 0.563& 0.602& 0.616& 0.617& 0.51& 0.581& \hl{0.639}& 0.402& 0.6\\
        %ItalyPowerDemand& 0.955& 0.948& 0.926& 0.953& 0.866& 0.951& \hl{0.97}& \hl{0.97}& \hl{0.97}\\
        LargeKitApp& 0.493& 0.823& 0.551& \hl{0.933}& 0.837& 0.816& 0.9& 0.896& 0.915\\
        Lighting2& 0.754& 0.71& 0.76& 0.659& 0.81& 0.835& 0.785& 0.813& \hl{0.869}\\
        %Lighting7& 0.575& 0.671& 0.68& 0.724& 0.666& 0.763& 0.799& \hl{0.863}& \hl{0.863}\\
        Mallat& 0.914& 0.929& 0.951& 0.972& 0.949& 0.961& 0.974& \hl{0.98}& 0.959\\
        %Meat& 0.933& \hl{0.983}& \hl{0.983}& 0.966& 0.98& 0.978& 0.981& 0.967& \hl{0.983}\\
        %MedicalImages& 0.684& 0.701& 0.707& 0.691& 0.715& 0.761& 0.785& 0.792& \hl{0.82}\\
        %MidPhxAgeGp& 0.519& 0.798& 0.8& \hl{0.815}& 0.808& 0.782& 0.801& 0.768& \hl{0.815}\\
        %MidPhxCorr& 0.766& 0.581& 0.673& 0.694& 0.666& 0.609& 0.722& 0.795& \hl{0.838}\\
        %MidPhxTW& 0.513& 0.519& 0.568& 0.579& 0.537& 0.525& 0.587& 0.612& \hl{0.634}\\
        %MoteStrain& 0.879& 0.891& 0.886& 0.882& 0.846& 0.857& 0.902& \hl{0.95}& 0.937\\
        NonECG1&  0.829& 0.877& 0.842& 0.947& 0.841& 0.849& 0.929& \hl{0.961}& 0.959\\
        NonECG2& 0.88& 0.898& 0.862& \hl{0.954}& 0.904& 0.914& 0.946& 0.935& 0.947\\
        OliveOil& 0.866& 0.864& 0.864& 0.881& 0.87& 0.879& 0.901& 0.833& \hl{0.903}\\
        %OSULeaf& 0.522& 0.78& 0.678& 0.934& 0.967& 0.812& 0.949& 0.988& \hl{0.992}\\
        %PhalangesOutlinesCorrect& 0.761& 0.793& 0.825& 0.794& 0.821& 0.78& 0.783& 0.826& \hl{0.836}\\
        Phoneme& 0.119& 0.22& 0.278& 0.329& 0.256& 0.299& \hl{0.362}& 0.345& 0.318\\
        %Plane& 0.963& 0.996& 0.993& \hl{1}& 0.998& \hl{1}& \hl{1}& \hl{1}& \hl{1}\\
        %ProximalPhalanxOAG& 0.785& 0.829& 0.861& 0.881& 0.867& 0.839& 0.871& 0.849& \hl{0.888}\\
        %ProximalPhalanxOC& 0.817& 0.824& 0.842& 0.841& 0.819& 0.805& 0.848& 0.9& \hl{0.931}\\
        %ProximalPhalanxTW& 0.713& 0.774& 0.798& 0.803& 0.773& 0.759& 0.815& 0.81& \hl{0.835}\\
        RefDevices& 0.395& 0.656& 0.638& 0.761& \hl{0.785}& 0.676& 0.742& 0.533& 0.589\\
        ScreenType& 0.56& 0.499& 0.538& 0.676& 0.586& 0.554& 0.651& 0.667& \hl{0.685}\\
        %ShapeletSim& 0.54& 0.888& 0.913& 0.934& \hl{1}& 0.827& 0.964& 0.867& \hl{1}\\
        ShapesAll& 0.753& 0.796& 0.853& 0.854& 0.909& 0.886& \hl{0.911}& 0.898& 0.892\\
        SmallKitApp& 0.342& 0.753& 0.674& 0.802& 0.75& 0.703& 0.788& 0.803& \hl{0.813}\\
        %SonyAIBORobotSurfaceI& 0.695& 0.884& 0.839& 0.888& 0.897& 0.794& 0.899& \hl{0.968}& 0.965\\
        %SonyAIBORobotSurfaceII& 0.859& 0.859& 0.825& 0.924& 0.888& 0.87& 0.96& \hl{0.962}& 0.949\\
        StarLCurves& 0.849& 0.96& 0.978& 0.977& 0.978& 0.941& \hl{0.98}& 0.967& 0.976\\
        Strawberry& 0.946& 0.97& 0.968& 0.968& 0.97& 0.959& 0.963& 0.969& \hl{0.987}\\
        %SwedishLeaf& 0.799& 0.885& 0.908& 0.939& 0.918& 0.916& 0.967& 0.966& \hl{0.971}\\
        %Symbols& 0.9& 0.93& 0.944& 0.862& 0.961& 0.957& 0.953& \hl{0.962}& 0.884\\
        %Synth Cntr& 0.88& 0.968& 0.987& 0.987& 0.968& 0.994& \hl{0.999}& 0.99& 0.997\\
        %ToeSegmentation1& 0.68& 0.922& 0.858& 0.954& 0.929& 0.788& 0.934& \hl{0.969}& 0.956\\
        %ToeSegmentation2& 0.808& 0.904& 0.886& 0.947& \hl{0.96}& 0.907& 0.951& 0.951& 0.931\\
        %Trace& 0.76& 0.997& 0.981& \hl{1}& \hl{1}& 0.996& \hl{1}& \hl{1}& \hl{1}\\
        %TwoLeadECG& 0.748& 0.958& 0.91& 0.984& 0.985& 0.985& 0.983& \hl{1}& \hl{1}\\
        %Two Patterns& 0.918& \hl{1}& 0.974& 0.952& 0.991& \hl{1}& \hl{1}& 0.897& \hl{1}\\
        uWavGestAll& 0.948& 0.963& 0.834& 0.942& 0.944& \hl{0.968}& 0.965& 0.826& \hl{0.968}\\
        %uWaveGestX& 0.739& 0.806& 0.746& 0.806& 0.753& 0.805& 0.831& 0.754& \hl{0.842}\\
        %uWaveGestY& 0.662& 0.717& 0.776& 0.737& 0.661& 0.731& \hl{0.766}& 0.725& 0.754\\
        %uWaveGestZ& 0.65& 0.736& 0.776& 0.747& 0.695& 0.726& 0.76& 0.729& \hl{0.788}\\
        %Wafer& 0.996& 0.996& 0.996& \hl{1}& 0.999& 0.997& 0.999& 0.997& 0.998\\
        %Wine&  0.611& 0.892& 0.879& \hl{0.926}& 0.912& 0.887& 0.904& 0.889& 0.889\\
        %WordsSynonyms& 0.618& 0.674& 0.669& 0.582& 0.659& \hl{0.778}& 0.748& 0.58& 0.666\\
        Worms& 0.454& 0.673& 0.668& 0.719& \hl{0.735}& 0.664& 0.725& 0.669& 0.696\\
        WormsTwoClass& 0.61& 0.73& 0.755& 0.779& 0.81& 0.717& 0.785& 0.729& \hl{0.796}\\
        %Yoga& 0.83& 0.863& 0.835& 0.823& \hl{0.91}& 0.885& 0.898& 0.845& 0.886\\
        Wins& 0& 0& 0& 4& 3& 1& 4& 3& \hl{10}\\
        AR& 8.2727& 6.0909& 6.8181& 3.3181& 4.4545& 5.6818& 2.7727& 4.0909& \hl{2.7272}\\
        GR& 8.1337& 5.7958& 6.6045& 2.8083& 3.8706& 5.1643& 2.4149& 3.3173& \hl{2.1653}\\
        MPCE& 0.1037& 0.0702& 0.0773& 0.0553& 0.0572& 0.0732& 0.0537& 0.0607& \hl{0.0501}\\
        \bottomrule
    \end{tabular}
\end{table*}

\begin{figure*}[t]
    \begin{minipage}{1.0\linewidth}
        \centering
        \centerline{\includegraphics[width=14cm]{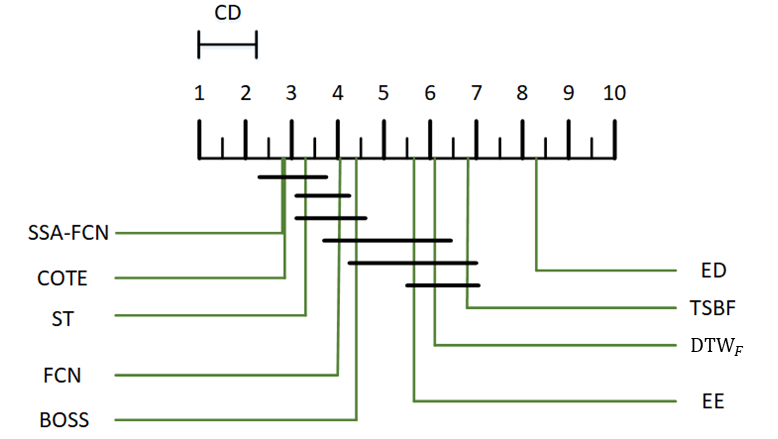}}
        %\vspace{2.0cm}
    \end{minipage}
    \caption{Critical difference diagram for the mean ranks of SARCoN and the other comparison algorithms. }
    \label{fig5:ar_rank}
\end{figure*}

\begin{figure*}[t]
    \begin{minipage}{1.0\linewidth}
        \centering
        \centerline{\includegraphics[width=16.5cm]{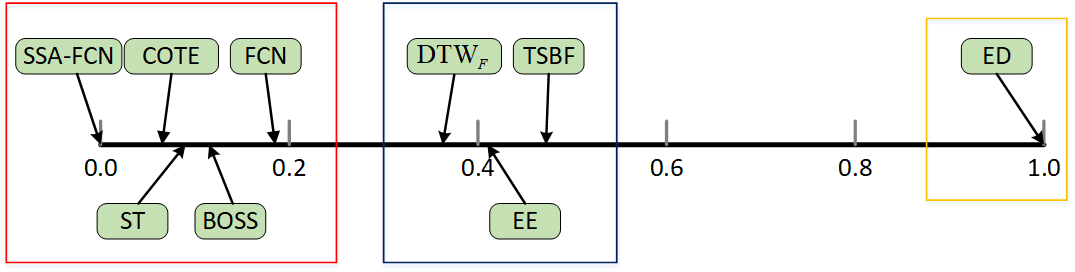}}
        %\vspace{2.0cm}
    \end{minipage}
    \caption{The paired T-test of the MPCE scores classifies models into three groups. The numerical values on the coordinate axis indicate the normalized MPCE scores.}
    \label{fig6:mpce_rank}
\end{figure*}

\subsection{Results}

Table \ref{tab:result} lists the classification accuracies of all the compared algorithms for the 24 UCR datasets. For each dataset, we ranked all nine classifiers from 1 to 9 in order of classification accuracy (the highest accuracy was given the rank of 1 and the lowest accuracy was given the rank of 9). The algorithms with the same performance for a given dataset were assigned the same ranking. The last four rows of the table show the number of wins, the rank-based statistics (AR and GR), and the MPCE of each algorithm for all datasets. The best result for each dataset is marked with a shadow. Our proposed method, SARCoN, outperformed the state-of-the-art classifiers on nine of the 24 datasets. For the rest of the datasets, SARCoN achieved results comparable to what the other algorithms achieved. Moreover, SARCoN attained the best performance according to the other three evaluation criteria, namely AR, GR, and MPCE. Specifically, SARCoN and ED achieved the best rank score and worst MPCE score, respectively. SARCoN outperformed the ED baseline on all the datasets, and it was also superior to the feature-based methods on most of the datasets. The ensemble-based methods, especially COTE, demonstrated competitive performance to a certain extent on select datasets.

To verify the soundness of our conclusion on the rank of the compared algorithms, we conducted two more statistical significance tests on the performances of all nine algorithms. The Friedman tests \cite{demvsar2006statistical,chen2009probabilistic,chen2013efficient} were first conducted, and they indicated that the differences in the performances of all nine algorithms on all 24 datasets were significant (i.e., the differences in the performances of the nine algorithms were not caused by chance). Furthermore, the post-hoc pairwise Nemenyi test was employed to evaluate the significance of the performance differences, where the significance level was set to 0.05 and the calculated critical difference value was 2.452. Figure \ref{fig5:ar_rank} shows the critical difference diagram over the mean ranks of all nine algorithms, where the thick horizontal lines indicate critical difference statistics. From the statistical results, we can draw the conclusion that SARCoN achieved the best performance in terms of average rank, and that it was clearly superior to the feature-based and distance-based methods such as BOSS, TSBF, $\textup{DTW}_F$, and ED. Among all the algorithms compared, COTE was the only one that performed comparably to SARCoN. Based on these results, we can conclude that SARCoN, as a self-contained algorithm without heavyweight feature engineering, obtains state-of-the-art performance for TSC problems according to most benchmarks.

The drawback of the rank-based criteria is their sensitivity to the selection of the model pool and its size. The MPCE is a more robust baseline criterion because it takes the number of classes into account. Therefore, paired T-tests were performed to evaluate the significance of MPCE differences between the algorithms. Figure \ref{fig6:mpce_rank} shows the results, where all algorithms are grouped into three categories based on the T-test results of the MPCE scores. The best group contained SARCoN, COTE, ST, FCN, and BOSS; $\textup{DTW}_F$, TSBF, and EE formed the second best group, and ED was included in the worst group. The differences between the MPCE scores within the same group were not significant.

Based on comprehensive statistical analyses of the experimental results, we can solidly conclude that ED performed the worst for long-term series. Although $\textup{DTW}_F$, TSBF, and EE belong to different algorithmic categories, they exhibited similar performances. By adopting a structured self-attention mechanism, SARCoN significantly improved the generalization performance of FCNs. Among all the algorithms, SARCoN and COTE demonstrated superior performance over all the other algorithms.

\begin{table*}[t]
    \centering
    %\caption{Details of the time and space complexity of each algorithm.}
    \caption{The details of the time and space complexity of the algorithms}
    \label{tab:t_s_com}
    \begin{tabular}{cccc}
        \toprule
        Algorithm& Time complexity& Space complexity& Parameter description\\
        \midrule
        ED& $O(m*n)$ &$O(n)$& $m$: series length, $n$: number of series\\
        $\textup{DTW}_F$& $O(n^2m^2)$& $O(n^2+n\alpha^l)$& $l$: word length\\
        TSBF& $O(rmnwlogn)$& $O(rm)$& $r$: number of trees, w: number of subseries\\
        ST& $O(n^2m^4)$& $O(kn)$& $k$: number of shapelets\\
        BOSS& $O(nm(n-w))$& $O(n\alpha^l)$& w: window length\\
        EE& $O(n^2m^4)$ & $O(m^2)$& \\
        COTE& $O(n^2m^4)$ & $O(knm^2)$&\\
        FCN& $O\left( {\sum\limits_{l = 1}^D {m{k_{wid}}{k_{len}}{c_{l - 1}}{c_l}} } \right)$& $O\left( {\sum\limits_{l = 1}^D {{k_{wid}}{k_{len}}{c_{l - 1}} + \sum\limits_{l = 1}^D {m{c_l}} } } \right)$& \tabincell{c}{$k_{wid}, k_{len}$: convolution kernel size, $c$: number \\of channels, $D$: number of convolution layers}\\
        \bottomrule
    \end{tabular}
\end{table*}
\begin{figure*}
    \centering
    \subfigure[class activation map]{
        \centering
        \includegraphics[width=8.7cm,height=6.7cm]{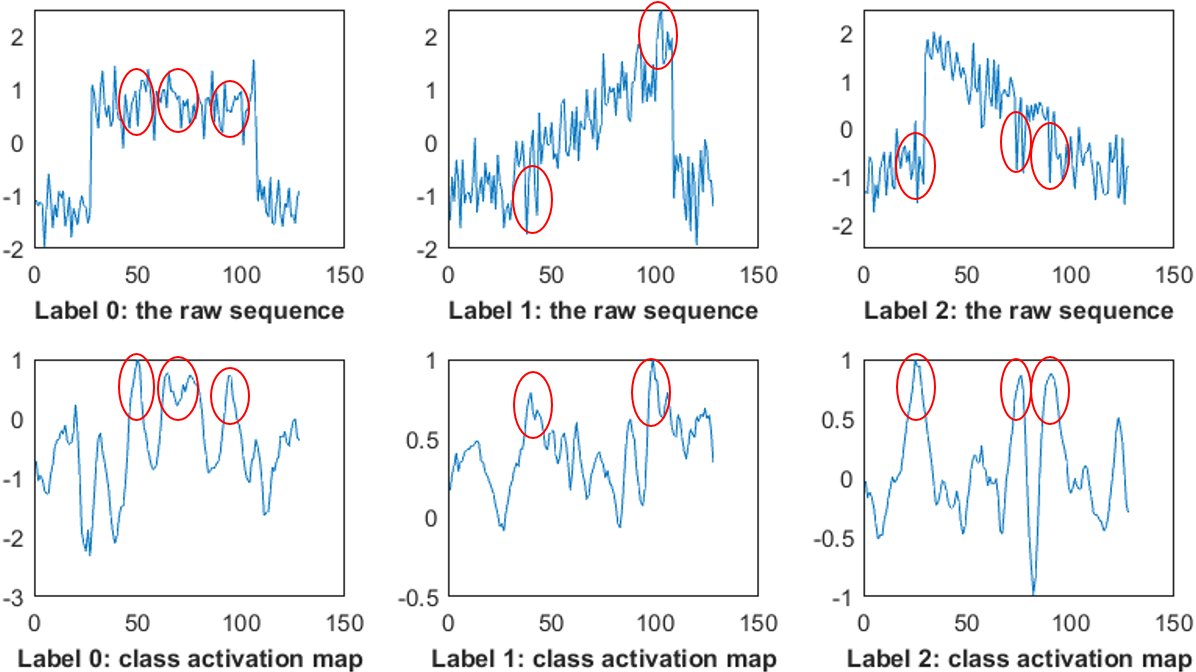}
        %\caption{average attention values}
    }
    \subfigure[attention values]{
        \centering
        \includegraphics[width=8.7cm,height=6.7cm]{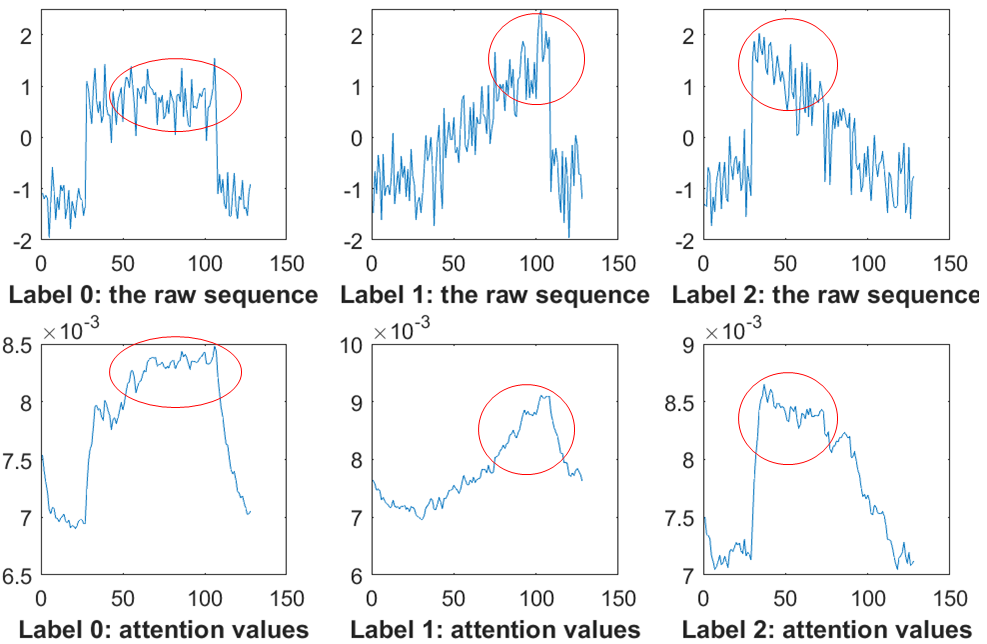}
        %\caption{class activation map}
    }
    \caption{Visualization of the contributing regions. The top row in (a) and (b) indicates the raw sequence of the three categories (0, 1, and 2). (a) The class activation map produced by the FCN block, and (b) the attention values learned by the SSA block.}
    \label{fig7:fcn_ssa bolck}
\end{figure*}

\subsection{Time and Space Complexity}
Although our main focus was on the classification accuracy of the algorithms, the time consumption of the algorithms is also worthy of consideration. We did not perform timing experiments to compare all the algorithms because we needed to ensure that the algorithm and codes we reproduced were optimized. Otherwise, bias and other unexpected consequences could have occurred. In addition, we also did not have an appropriate method for comparing the runtimes of all the algorithms. For instance, the ED algorithm has no training time, while FCNs and SARCoN are deep neural network models for which the evaluation indicator is usually the average training time per epoch. Thus, the use of this indicator would have rendered the comparison of the training time between algorithms meaningless. However, as there were large differences between the runtimes of and memory consumed by each algorithm, the most meaningful analysis was to compare the complexities, especially the space complexity, of the algorithms. Table \ref{tab:t_s_com} lists the time and space complexities of the algorithms. For the FCN, the time complexity was mainly subject to the convolutional layer, while the space complexity contained two parts: the total parameter quantity and the feature map of each layer. For SARCoN, the LSTM phase controlled the time complexity of the SSA submodule ($O(m)$), and the time complexity of SARCoN arose from the maximum time complexity of the SSA and FCN submodules. Apparently, extra space complexity came from the SSA submodule, including parameters from the LSTM ($O(p^2)$), attention matrix $A$  ($O(mt)$), sequence embedding $Q$ ($O(pt)$), weight matrices $W^1$ and $W^2$ ($O(ps+st)$), and parameters from the fully connected layers ($O(prh)$). Thus, the total space complexity of the SSA submodule was $O(p^2+(m+p)t+(p+t)s+prh)$, where $p$ is the size of the hidden state from LSTM and $h$ indicates the number of neurons in the second fully connected layer.

To demonstrate the time consumption of SARCoN and the FCN for specific tasks, extended experiments were conducted on six benchmark time series datasets. The classification model ran on the server, and the GPU type was GeForce RTX 2080. The average training time per epoch was used as the time standard. Table \ref{tab:time_cons} shows the comparison results for the training times and classification accuracies.

\begin{table}[t]
    \centering
    %\caption{Average training time per epoch for SARCoN and the FCN using the benchmark time series datasets.}
    \caption{The average training time of each epoch for SARCoN and FCN on time series benchmarks}
    \label{tab:time_cons}
    \begin{tabular}{|c|c|c|c|c|}
        \hline
        \multirow{2}{*}{Datasets}& \multicolumn{2}{c|}{FCN} &\multicolumn{2}{c|}{SARCoN}\\
        \cline{2-5}
        \multicolumn{1}{|c|}{}&time(s)&accuracy&time(s)&accuracy\\
        \hline
        Computers& 0.132& 0.848& 0.211& 0.868\\
        \hline
        RefDevices& 0.138& 0.533& 6.48& 0.589\\
        \hline
        Haptics& 0.152& 0.551& 3.784& 0.526\\
        \hline
        Herring& 0.436& 0.703& 0.768& 0.734\\
        \hline
        Worms& 0.437& 0.669& 3.057& 0.696\\
        \hline
        ScreenType& 1.34& 0.667& 7.35& 0.685\\
        \hline
    \end{tabular}
\end{table}

Several conclusions were drawn from the experimental results: (1) the ED algorithm had the lowest time and space complexity; (2) the feature-based (TSBF, ST, and BOSS) and ensemble-based (EE and COTE) approaches had higher time and space complexity; (3) the improvement in the classification performance of SARCoN compared to the FCN was achieved at the cost of increases in time and space complexity, but to an acceptable degree.

%\begin{figure}[t]
%   \setlength{\belowcaptionskip}{-0.5cm}
%   \begin{minipage}{1.0\linewidth}
%       \centering
%       \centerline{\includegraphics[width=9cm]{attention_visual.pdf}}
%       %\vspace{2.0cm}
%   \end{minipage}
%   \caption{Visualization of average attention values on CBF dataset.}
%   \label{fig7:av}
%\end{figure}

\subsection{Visualization of Contributing Regions}
In this subsection, we explore the extent to which the FCN and SSA submodules contribute to the classification decision of our proposed model by visualizing the corresponding discriminative areas of the submodules.

%To what extent does the FCN and SSA sub-module contribute to the classification decision of our proposed model, respectively? In this subsection, we will explore the contributions to the discriminative areas that determine the label of the time series to reveal of the corresponding regions of the FCN sub-module and the SSA sub-module for the time series.

The Class Activation Map (CAM) was proposed by \cite{zhou2016learning} to reveal the discriminative image regions for a particular category based on convolutional neural networks. Here, we employ CAM to discover the contributing regions of the time series controlled by the FCN block. For a classified time series, let $f_k\left( t \right)$ represent the output of the filter $k$ in the third basic block of the FCN at time step $t$. Then, for the filter $k$, the result of the global average pooling layer is ${F^k} = \sum\limits_t {{f_k}\left( t \right)}$. For the specific class $c$ of the softmax layer, $w_k^c$ is the corresponding weight of class $c$ for filter $k$. Hence, the input to the softmax layer is

\begin{equation}
{S_c} = \sum\limits_k {w_k^c\sum\limits_t {{f_k}\left( t \right)} } \\
= \sum\limits_k {\sum\limits_t {w_k^c{f_k}\left( t \right)} }.
\end{equation}

Let ${I_c}$ indicate the CAM for class $c$ at time step $t$. Then,
\begin{equation}
{I_c\left(t,c\right)} = \sum\limits_k {w_k^c{f_k}\left( t \right)}.
\end{equation}

The variable $I_c \left(t,c\right)$ directly reflects the importance of the activation at time step $t$, as it leads to the classification results of the time series. As the length of the output of the global average pooling layer was different from that of the input, the class activation map needed to be upsampled to the same length as the input time series to clearly demonstrate the contributing regions of different submodules in the input time series.

To understand and visualize the structure of the decision process of the SSA submodule, a second intelligible method proposed by \cite{lin2017structured} was adopted. Specifically, an attention vector was obtained by summing all attention vectors in $A$ and normalizing the resulting attention vector to sum up to 1. Each element value in the vector reflected the importance of the corresponding temporal location in the time series.

We then conducted a set of experiments on the `CBF’ dataset using the analysis methods mentioned above. The visual experimental results are depicted in Figure \ref{fig7:fcn_ssa bolck}. The top row in (a) and (b) indicate the raw sequence of the three categories (0, 1, and 2). The second rows in (a) and (b) contain the corresponding class activation maps and attention values from the forward process, respectively. The results further confirm the contribution of each submodule: several narrow single peaks in the CAM can be readily spotted, which is strong evidence that the FCN was able to extract local information. In contrast, the behavior of the SSA was very different from the FCN, as the longer continuous segments detected by the SSA can be easily observed, which indicates that the SSA can extract long-term pattern information.

Specific category diagrams can generate further insight into how the FCN and SSA submodules contribute to the classification decision. For label 0, the SSA extracted certain segments with important information that the FCN did not capture. However, for labels 1 and 2, the FCN discovered some important information that the SSA did not. Therefore, we can infer from the experimental results that the FCN focuses more on local regions, while the SSA focuses more on the trend of the whole sequence, even though they work on the same region of attention. We can even postulate that it is the diversity in the roles of the two submodules that enables our model to outperform other models on certain datasets.

% Of course, they both pay attention to some important information in the same regions. We can get an overall intuitive understanding that.

There are additional conclusions that can be drawn from Figure \ref{fig7:fcn_ssa bolck}. Sequences with label 0 are distinguished mostly by the high and flat regions in the middle of the sequence. For label 1, the discriminative information is mainly distributed in the region where the sequence first rises smoothly and then drops sharply. The region where the sequence first rises sharply and then drops smoothly determines where label 2 is applied.

\begin{figure}[t]
    \centering
    \includegraphics[width=8.5cm]{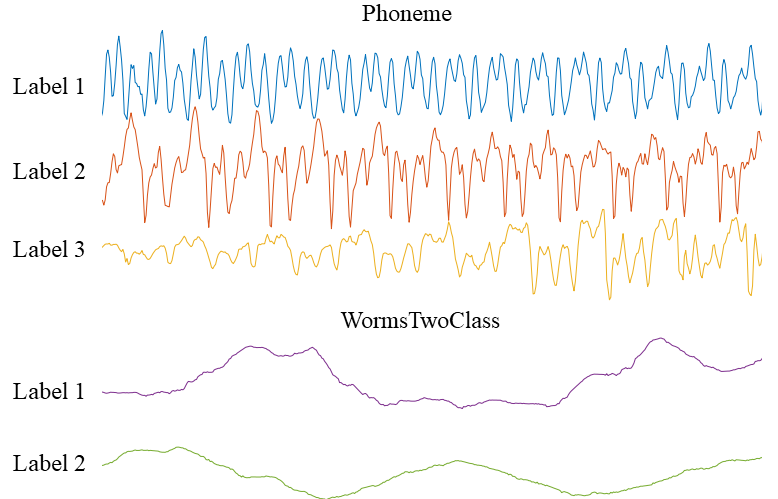}
    \caption{Time series with different labels from Phoneme and WormsTwoClass.}
    \label{fig9:short_long}
\end{figure}
\begin{figure*}[t]
    \centering
    \includegraphics[width=17cm]{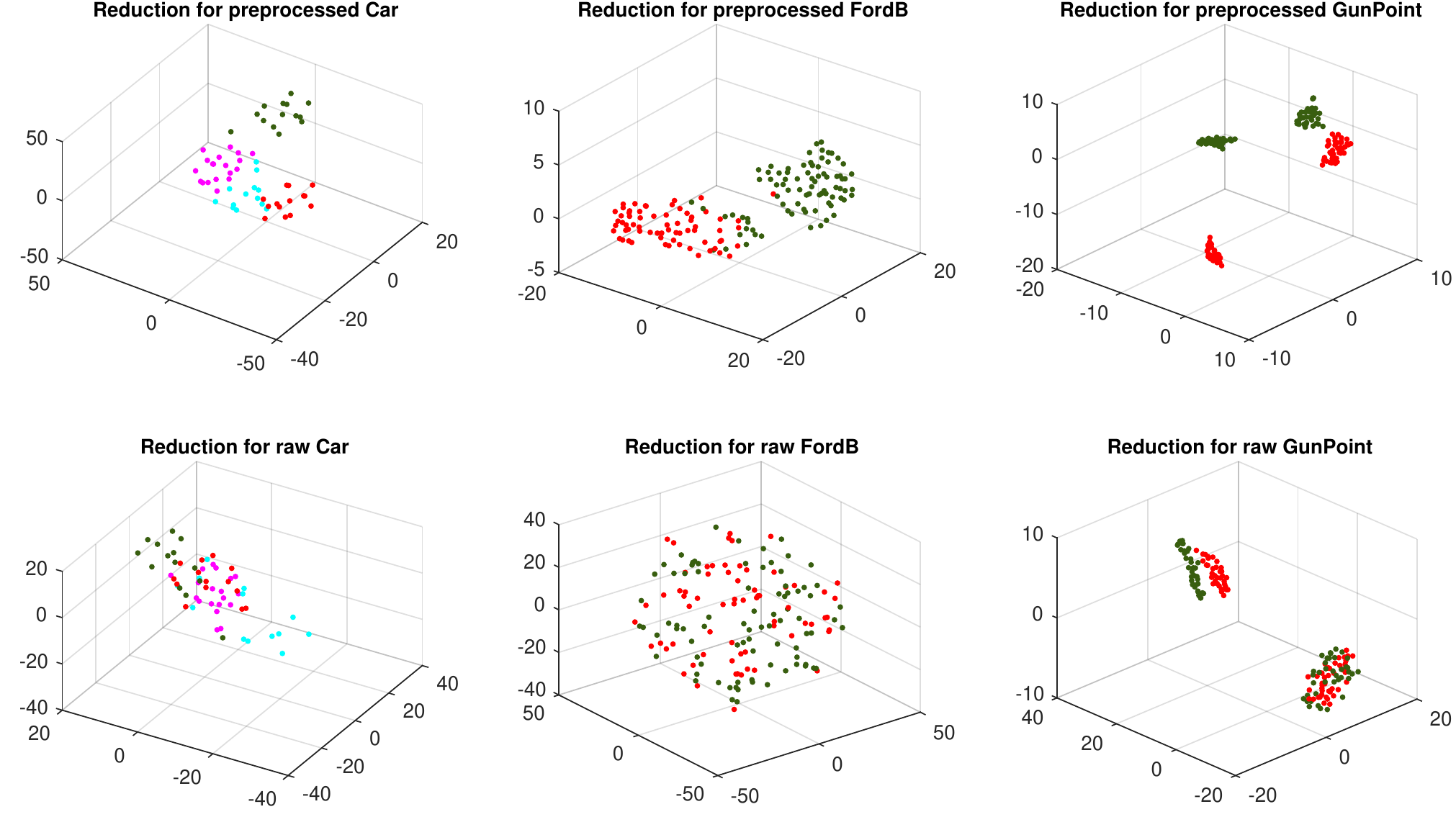}
    \caption{Visualization of the raw data and the data extracted by T-SNE on time series benchmarks. The top row displays the results of the dimension reduction of the features extracted by our model. The following row shows the results of the dimension reduction of the original sequence.}
    \label{fig8:tsne_feature}
\end{figure*}

\subsection{Model Applicability Discussion}
Although the experimental results in Table \ref{tab:result} demonstrate that SARCoN was superior to the FCN for most of the datasets (e.g., WormsTwoClass and Computers), it did not perform as well as expected on other datasets (e.g., Phoneme and Haptics). In this subsection, we use the Phoneme and WormsTwoClass benchmarks as examples to discuss the applicability of our proposed model. Figure \ref{fig9:short_long} depicts the time series in Phoneme and WormsTwoClass with different labels. For the Phoneme benchmark, each series was extracted from the segmented audio, which were sampled from male and female speakers with different ratios. Periodic local feature patterns can be easily observed in the sequences. For the WormsTwoClass benchmark, each series was derived from the trajectory of the worms, where long-term temporal dependency information was contained in the sequences.

\begin{table}[t]
    \centering
    %\caption{Comparison between classification accuracies of the FCN and SARCoN for the Phoneme and WormsTwoClass benchmarks.}
    \caption{Comparison of classification accuracy between FCN and SARCoN on Phoneme and WormsTwoClass benchmarks}
    \label{tab:fcn_ssa_fcn}
    \scalebox{1.0}{
        \begin{tabular}{p{2.6cm}<{\centering}p{2cm}<{\centering}p{2cm}<{\centering}}
            \toprule
            Datasets& FCN& SARCoN\\
            \midrule
            Phoneme& 0.345& 0.318\\
            WormsTwoClass& 0.729& 0.796\\
            \bottomrule
    \end{tabular}
}
\end{table}

Figure \ref{tab:fcn_ssa_fcn} presents the corresponding experimental results, from which we can observe that SARCoN performed better than FCN on the WormsTwoClass dataset, whereas FCN was superior to SARCoN on the Phoneme dataset. These experimental results reveal that the FCN is better at processing time series featuring local discriminative patterns, but SARCoN is more suitable for time series with long-term temporal dependencies. We posit the reason for this fact is that the amalgamation of long-distance representations compromises the discriminative power of local temporal patterns, and SARCoN’s improvement in generalization is the result of the balance between its local and the long-distance temporal representation abilities.

%These observations are consistent with the design rational of each submodule for feature extraction.

\subsection{Visualization on Reduced Dimensional Space}

Additional experiments demonstrated how the features from the penultimate layer (i.e., `concat’) of our proposed method contributed to the classification performance. We conducted the experiments on three time series benchmarks, including Car, FordB, and GunPoint. The t-distributed stochastic neighbor embedding (T-SNE) \cite{maaten2014accelerating}, an effective technology for dimensionality reduction, was adopted for visualizing high-dimensional datasets. In the experiments, three important parameters were provided to the `tsne’ function in the Matlab toolkit: the data matrix $X$, algorithmic type, and number of dimensions. Other parameters were set using the default settings. The T-SNE was used to embed the original high-dimensional data into three-dimensional space. In the figure, the sequences with the same labels are colored the same, and the sequences with different labels are distinguished by different colors. We rotated the resulting projection to the most informative perspective from which the grouping patterns could be readily observed, as shown in Figure \ref{fig8:tsne_feature}.

From the experimental results, we can observe that, compared to the original data, the extracted features with the same label were closer together, and the ones with different labels were more distant from each other. In other words, in the embedded space, the distances between data within the same class were much smaller than the distances between data dispersed in different classes. This experimental observation verifies that SARCoN is capable of extracting valid discriminative features in TSC problems.

\section{Conclusions}
\label{sec:conclu}

In this study, we elaborated a novel hybrid DNN architecture in which the FCN submodule and self-attentive LSTM submodule were used to learn multi-faceted representations from time series. The FCN submodule was dedicated to capturing the hierarchical local temporal features of the input time series, and the self-attentive LSTM submodule was used to learn long-distance dependencies in the input time series. The different submodules were trained jointly in an end-to-end manner, whereas for the prediction of unseen data, the features with different perspectives were extracted in parallel and concatenated to make the classification decision using a holistic approach. Our experiments on the UCR benchmark datasets showed that SARCoN achieved a performance comparable to that of conventional state-of-the-art approaches, but with considerably fewer complexities. To gain insight into the reason for SARCoN’s significant improvement in performance over the other algorithms, we conducted a series of well-defined empirical analyses. We draw the following conclusions from our work:

\begin{enumerate}
    \item Multi-faceted representations of time series that are learned from different perspectives play an essential role in TSC. Time series usually contain complicated correlations at different scales over time, and thus one-sided representations lead to the loss of critical discriminative temporal information. The amalgamation of local temporal representations with long-distance representations can effectively improve the expressive power of time series.
    \item  Although currently there is no single neural structure that can learn the local and long-distance temporal correlations of time series equally well in the same amount of time, a hybrid neural architecture that integrates various neural structures provides an effective means to learn holistic representations with acceptable increases in time and space complexity. As such a hybrid architecture, SARCoN significantly improves the overall generalization performance for TSC.
\end{enumerate}

There are several future research directions that extend from this work. First, the analysis of multivariate time series is a natural extension of the proposed architecture, and thus we will extend the proposed approach to accommodate this additional analysis. However, a more intelligent way to construct the neural architecture and assemble the holistic representations is required to address the high-dimensional problem of multivariate time series. Second, the performance of the proposed model is limited by the amount of training data because it was trained with supervised means. Due to the sparsity of carefully labeled time series data, it is worth exploring unsupervised learning to initially train the model with a much larger amount of unlabeled data, and then fine-tune the model with less labeled data using supervised learning. Finally, we will continually explore the mechanism behind the trade-off between the local and long-distance representations that was revealed by our experiments. Based on a deeper understanding of this trade-off, we expect to find a systematic solution to the mutual increase in the local discrimination and long-distance recognition in order to further enhance the generalization ability of the TSC models.

\bibliographystyle{IEEEtran}
\bibliography{reference}
\end{document}